\documentclass[letterpaper, 10pt, journal, twoside]{IEEEtran}

\usepackage[utf8]{inputenc}
\usepackage{amssymb,amsfonts,amsmath,amscd}
\usepackage{bm}
\usepackage[pdftex]{graphicx}
\usepackage{url}
\usepackage{cite}
\usepackage[usenames,dvipsnames]{color}
\usepackage{subcaption}
\usepackage{caption}
\usepackage{accents}
\usepackage{tabularx}
\usepackage{hyperref}

\usepackage[export]{adjustbox}

\usepackage{enumitem}

\usepackage{mathtools}
\usepackage[english]{babel}
\usepackage{siunitx}
\usepackage[export]{adjustbox}

\usepackage{microtype}

\usepackage{booktabs}
\usepackage{makecell}
\usepackage{multirow}

\definecolor{pltred}{rgb}{0.839, 0.153, 0.157}

\DeclareMathOperator*{\argmin}{argmin}

\newcommand{\R}{\mathbb{R}}

\newcommand{\cmd}{\mathrm{cmd}}

\newcommand{\videourl}{http://tiny.cc/force-push}
\newcommand{\codeurl}{https://github.com/utiasDSL/force\_push}

\title{Force Push: Robust Single-Point Pushing with Force Feedback}
\author{Adam Heins and Angela P. Schoellig%
\thanks{This article was published in IEEE Robotics and Automation Letters. Digital Object Identifier (DOI): 10.1109/LRA.2024.3414180}%
\thanks{This work was supported by the Natural Sciences and Engineering Research Council of Canada and the Canadian Institute for Advanced Research.}%
  \thanks{The authors are with the Learning Systems and Robotics Lab (www.learnsyslab.org) at the Technical University of Munich, Germany, and the University of Toronto Institute for Aerospace Studies, Canada. They are also affiliated with the University of Toronto Robotics Institute, the Munich Institute of Robotics and Machine Intelligence (MIRMI), and the Vector Institute for Artificial Intelligence. E-mail: adam.heins@robotics.utias.utoronto.ca, angela.schoellig@tum.de}%
  \thanks{\copyright\ 2024 IEEE.  Personal use of this material is permitted.  Permission from IEEE must be obtained for all other uses, in any current or future media, including reprinting/republishing this material for advertising or promotional purposes, creating new collective works, for resale or redistribution to servers or lists, or reuse of any copyrighted component of this work in other works.}%
}

\begin{document}

\maketitle

\begin{abstract}
  We present a controller for quasistatic robotic planar pushing with
  single-point contact using \emph{only} force feedback to sense the
  pushed object. We consider an omnidirectional mobile robot pushing an object
  (the ``slider'') along a given path, where the robot is equipped with a
  force-torque sensor to measure the force at the contact point with the
  slider. The geometric, inertial, and frictional parameters of the slider are
  not known to the controller, nor are measurements of the slider's pose. We
  assume that the robot can be localized so that the global position of the
  contact point is always known and that the approximate initial position of
  the slider is provided. Simulations and real-world experiments show that our
  controller yields pushes that are robust to a wide range of slider parameters
  and state perturbations along both straight and curved paths. Furthermore, we
  use an admittance controller to adjust the pushing velocity based on the
  measured force when the slider contacts obstacles like walls.
\end{abstract}

\section{Introduction}

Pushing is a nonprehensile manipulation primitive that allows
robots to move objects without grasping them, which is useful for objects that
are too large or heavy to be reliably grasped~\cite{stuber2020lets}. In this
work we investigate robotic planar pushing with single-point contact using only
force feedback to sense the pushed object (``the slider''), in contrast to
previous works on single-point pushing, which use visual or tactile sensing;
e.g.,~\cite{emery2001behavior,igarashi2010a,lynch1992manipulation,okawa1992control}.
The ``pusher'' is an omnidirectional mobile robot equipped with a force-torque
(FT) sensor to measure the contact force applied by the robot's end effector
(EE) on the slider through the contact point. The geometric, inertial (i.e.,
mass, center of mass (CoM), and inertia), and frictional parameters of the
slider are assumed to be unknown. We also assume that online feedback of the
slider's pose is not available---the only measurement of the slider is through
the contact force. Finally, we assume that the global position of the pusher is
known at all times (i.e., the robot can be localized) and that motion is
quasistatic.

Single-point pushing is a simple approach that does not require assumptions
about the slider's shape. In particular, we envision our force-based approach
being useful for pushing unknown objects between distant waypoints, where
reliable localization of the object is not available (e.g., due to poor
lighting). For example, consider pushing objects through hallways within
warehouses or factories. In this case, we are not concerned about tracking a
path exactly at all times, but rather ultimately pushing the object from one
place to another. Our approach can handle collisions with obstacles like walls,
using an admittance controller to regulate the contact force. Furthermore, we
hope that the insights presented here for force-based pushing can be combined
with other sensing modalities to improve overall performance.

\begin{figure}[t]
  \centering
    \includegraphics[width=\columnwidth]{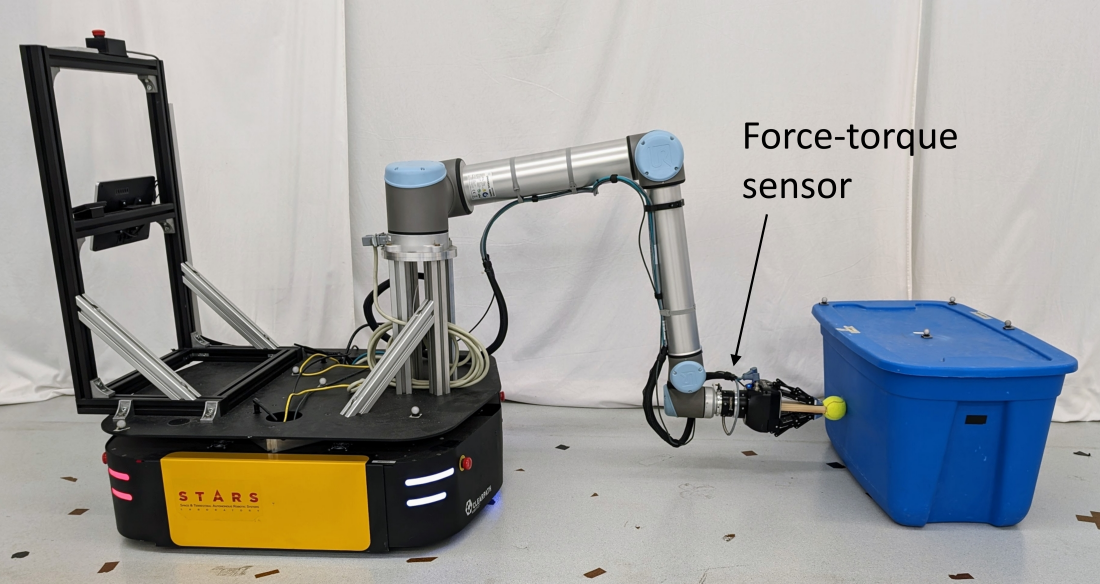}
    \caption{Our robot pushing a blue box across the floor using
      single-point contact. The contact force is measured by a force-torque
      sensor in the robot's wrist, but no other measurements of the object are
      provided. A video of our approach is available at
      \textbf{\texttt{\scriptsize \videourl}}.}
  \label{fig:eyecandy}
\end{figure}

The main contribution of this work is a controller for robotic single-point
pushing using \emph{only} force feedback to sense the pushed object. We show
that it successfully pushes objects along both straight and curved paths with
single-point contact and no model of the object. We demonstrate the robustness
of the controller by simulating pushes using a wide variety of slider
parameters and initial states. We also present real hardware experiments in
which a mobile manipulator successfully pushes different objects across a room
(see Fig.~\ref{fig:eyecandy}) along straight and curved paths, including some
with static obstacles. Notably, we do not assume that sufficient friction is
available to prevent slip at the contact point; slipping is a natural part of
the behaviour of our controller and does not necessarily lead to task failure.
Our code is available at \texttt{\small \codeurl}.

We first briefly described a controller for single-point pushing based
on force feedback in~\cite{heins2021mobile}, but we have substantially changed
and augmented this approach in the current work. In particular, we reformulate
the pushing control law, add a term to track a desired path, add
admittance control to handle obstacles, provide an analysis of its robustness
in simulation, and perform more numerous and challenging real-world
experiments.

\section{Related Work}

Research on robotic pushing began with Mason~\cite{mason1986mechanics}, and a
recent survey can be found in~\cite{stuber2020lets}. Early approaches were
typically either open-loop planning methods that rely on line contact for
stability~\cite{lynch1996stable,akella1998posing} or feedback-based approaches
using vision~\cite{emery2001behavior,igarashi2010a} or tactile
sensing~\cite{okawa1992control,lynch1992manipulation}. Tactile sensing is the
most similar to our work, though we assume only a single contact force vector
is available, rather than the contact angle and normal that a tactile sensor
provides. Furthermore,~\cite{lynch1992manipulation} only focuses on stable
translational pushes (i.e., where the slider moves in a constant direction)
without path-tracking, and~\cite{okawa1992control} assumes a model of the
slider is available and that there is sufficient friction to prevent slip.

An FT sensor is used with a fence to orient polygonal parts using a sequence of
one-dimensional pushes in~\cite{rusaw1999part}, which was shown to require less
pushes than the best sensorless alternative~\cite{akella2000parts}. Another use
of an FT sensor was in~\cite{ruizugalde2011fast}, where FT measurements are
used to detect slip while pushing. In contrast, we do not detect slip, and our
controller can still successfully push an object despite (unmeasured) slip.

In~\cite{nieuwenhuisen2005path}, a path planning method is proposed for pushing
an object while exploiting contact with obstacles in the environment. However,
while we rely on an admittance controller to adjust the commanded velocity
based on sensed force, the approach in~\cite{nieuwenhuisen2005path} exploits
the geometry of the obstacles to provide additional kinematic constraints on
the slider's motion. Furthermore, in~\cite{nieuwenhuisen2005path} the obstacles
are assumed to be frictionless and the approach is limited to a disk-shaped
slider, while we demonstrate multiple slider shapes in contact with obstacles
with non-zero friction.

More recent work uses supervised learning to obtain models of the complex
pushing dynamics arising from uncertain friction distributions and object
parameters. In~\cite{bauza2017a}, the pushing dynamics are learned using
Gaussian process (GP) regression. In~\cite{li2018push} the authors propose
Push-Net, an LSTM-based neural network architecture for pushing objects using
an image mask representing the slider's pose. In~\cite{kloss2022combining}, the
analytical model of quasistatic pushing from~\cite{lynch1992manipulation} is
combined with a learned model, which maps the slider position and depth images
to the inputs of the analytical model. Pushing is also a popular task for
reinforcement learning (RL). RL with dynamics randomization is used
in~\cite{peng2018sim} to train a pushing policy, which is then transferred to a
real robot arm with no fine-tuning. In~\cite{ferrandis2023nonprehensile}, a
multimodal RL policy is trained in simulation, which is hypothesized to better
represent the underlying hybrid dynamics of planar pushing compared to previous
unimodal policies. These supervised learning and RL works all rely on visual
feedback to obtain (some representation of) the object's pose, and are also
typically focused on pushing small objects on a tabletop.

Another recent line of work uses model predictive control (MPC) for fast online
replanning. The GP-based model in~\cite{bauza2017a} is used for MPC
in~\cite{bauza2018a}. In~\cite{hogan2020feedback}, hybrid contact dynamics
(with hybrid modes corresponding to sticking and sliding of the contact point)
are incorporated into the controller using approximate hybrid integer
programming. The approach based on mathematical programming with complementary
constraints in~\cite{moura2022non} is more computationally expensive
than~\cite{hogan2020feedback} but can handle complete loss of contact between
objects. In~\cite{agboh2020pushing}, the dynamics of pushing are modelled as a
Markov decision process, thus taking stochasticity into account. This allows
the controller to adjust its speed based on how much uncertainty can be
tolerated for each part of the task. These works all depend on feedback of the
slider's pose as well as a (learned or analytical) model of its dynamics. A
linear time-varying MPC approach and a nonlinear MPC approach are developed for
nonholonomic mobile robots pushing objects with line contact
in~\cite{bertoncelli2020linear} and~\cite{tang2023unwieldy}, respectively. Line
contact allows the controller to assume the slider stays rigidly attached to
the pusher as long as the friction cone constraints are satisfied, similar to
the open-loop planning approaches in~\cite{lynch1996stable}
and~\cite{akella1998posing}. However, line contact requires the slider to have
a flat edge to push against, which excludes, e.g., cylindrical sliders.

Finally, quadruped robots have also been used to push objects across the
ground~\cite{rigo2023contact} and up slopes~\cite{sombolestan2023hierarchical}
while regulating the required pushing force. In~\cite{rigo2023contact},
friction between the pusher and slider is neglected so that the resulting MPC
optimization problem is linear in both contact force and contact location, and
it is assumed that the slider's measured pose is available.
In~\cite{sombolestan2023hierarchical}, an adaptive MPC framework is used to
push objects with unknown (and possibly slowly varying) mass and friction
parameters; it is the only work that assumes the object model is unknown and
only interacts with it through the contact force, like we do. However,
in~\cite{sombolestan2023hierarchical} it is assumed that line contact between
pusher and slider is present, and curved paths are not considered. In contrast,
we consider single-point contact and show that our controller can push the
slider along curved paths.

In summary, most methods assume at least visual or tactile feedback of the
slider is available or assume line contact. Many also require an a priori model
of the slider and may be expensive to evaluate if the hybrid dynamics are taken
into account. \emph{None} of these methods perform single-point pushing using
\emph{only} force feedback to sense the slider.

\section{Problem Statement}

Our goal is to push an object along a given continuous planar
path~$\bm{p}_d(s)\colon\R_{\geq0}\to\R^2$, parameterized by the
distance~$s\geq0$ from its start. In this work, we use paths made up of
straight-line segments and circular arcs. We assume we have a
velocity-controlled pusher that is capable of measuring the planar
force~$\bm{f}\in\R^2$ it applies on the slider. Our controller must generate a
commanded velocity~$\bm{v}_{\mathrm{cmd}}\in\R^2$ for the pusher's EE at each
control timestep, which pushes the slider along the desired path.

We assume that the motion of the slider is quasistatic (i.e., inertial forces
are negligible), which means that the slider does not move when not in contact
with the pusher. We assume that the slider's geometric, inertial, and
frictional properties are unknown, except that its shape is convex. We also
assume that the slider is a single, non-articulated body (e.g., no wheels or
moving joints). Finally, we assume that an approximate initial position of the
slider is available, so that the robot can be positioned such that it makes
first contact with the slider by moving forward in the direction of the desired
path.

\section{Task-Space Pushing Controller}

\begin{figure}[t]
  \centering
    \includegraphics[width=0.7\columnwidth]{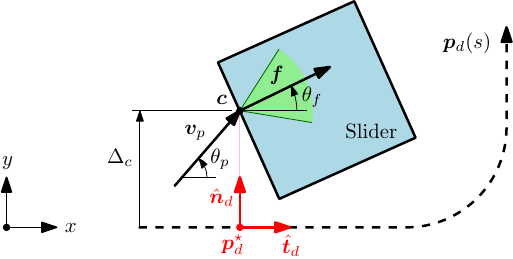}
    \caption{Example of our pushing controller with a square slider. The goal
      is to push the slider along the path~$\bm{p}_d(s)$ by pushing with
      velocity~$\bm{v}_p$ at the contact point~$\bm{c}$. The resulting contact
      force~$\bm{f}$ lies inside the friction cone (green). The pushing
      angle~$\theta_p$ is proportional to the lateral offset~$\Delta_c$ and
      difference between measured force angle and the desired path
      heading~$\Delta_f=\theta_f-\theta_d$; all angles are measured with
      respect to the global fixed frame. In this example, the pushing
      velocity~$\bm{v}_p$ will eventually rotate the slider so that the contact
      force points back toward the desired path. Depending on the contact
      friction coefficient~$\mu_c$, the contact point may slip along the
      slider's edge over the course of a trajectory.}
  \label{fig:controller}
\end{figure}

\begin{figure}[t]
  \centering
    \includegraphics[width=\columnwidth]{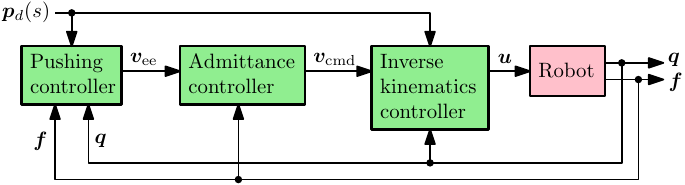}
    \caption{Block diagram of the system. The components of our controller (in
    green) use measurements of the robot's pose~$\bm{q}$ and contact
    force~$\bm{f}$ to produce joint velocity inputs~$\bm{u}$ that push the
    slider along a desired path~$\bm{p}_d(s)$.}
  \label{fig:block_diagram}
\end{figure}

When the properties of the slider are known, we can predict its motion using the
equations of motion given in~\cite{lynch1992manipulation}. However, since we do
not assume to know the geometry or inertial properties of the slider, we do not
rely on a particular mathematical motion model in our controller. Instead, we
use the following intuition---similar to Mason's Voting
Theorem~\cite{mason1986mechanics}---to control the system based on contact
force measurements. Suppose the object is starting to turn counterclockwise,
but we would like the robot to push it straight, as in
Fig.~\ref{fig:controller}. The contact force vector~$\bm{f}$ will also start
rotating counterclockwise. Then we need to push in a direction \emph{even
further} in the counterclockwise direction to eventually rotate the object back
clockwise and toward the straight-line direction. This controller is
essentially a proportional controller which acts on the pushing angle, except
that we actually need to turn further \emph{away} from the desired path in
order to correct the error. This yields a behaviour that trades-off short-term
error for long-term performance, which is more typically seen in predictive
controllers that consider the effect of their actions over a horizon.  A block
diagram of the entire controller is shown in Fig.~\ref{fig:block_diagram}; the
different components are described in the remainder of this and the following
section.

\subsection{Stable Pushing and Path-Tracking}

Let~$\bm{v}_p\in\R^2$ be the pushing velocity of the contact point, which,
together with the contact force~$\bm{f}$, we express in polar coordinates with
respect to the global frame as
\begin{align*}
  \bm{v}_p &= v\begin{bmatrix} \cos\theta_p \\ \sin\theta_p \end{bmatrix}, &
  \bm{f} &= \|\bm{f}\|\begin{bmatrix} \cos\theta_f \\ \sin\theta_f \end{bmatrix},
\end{align*}
where~$v\triangleq\|\bm{v}_p\|$ is the pushing speed. We will start by
taking~$v$ to be constant and controlling the pushing angle~$\theta_p$. We
denote unit vectors with~$\hat{(\cdot)}$, such that~$\skew5\hat{\bm{f}}$ is the
unit vector pointing in the direction of~$\bm{f}$. At the current timestep,
let~$\bm{p}_d^{\star}$ be the closest point on the desired path to the contact
point~$\bm{c}$ (in practice we assume~$\bm{c}$ is some fixed point on the
pusher's EE) and let us attach a Frenet-Serret frame at that point, with
directions~$\hat{\bm{t}}_d$ pointing tangent to (along) the path
and~$\hat{\bm{n}}_d$ orthogonal to the path (see Fig.~\ref{fig:controller}). We
denote the angle from the~$x$-axis to~$\hat{\bm{t}}_d$ as~$\theta_d$. Our
pushing control law is simply
\begin{equation}\label{eq:ctrl_law}
  \theta_p = \theta_d + (k_f+1)\Delta_f + k_c\Delta_c,
\end{equation}
where~$k_f,k_c>0$ are tunable gains, $\Delta_f=\theta_f-\theta_d$ is the signed
angle between~$\skew5\hat{\bm{f}}$ and~$\hat{\bm{t}}_d$,
and~$\Delta_c=\hat{\bm{n}}_d^T(\bm{c} - \bm{p}_d^{\star})$ is the lateral
offset from the path. The~$\Delta_f$ term steers toward a stable translational
pushing direction and the~$\Delta_c$ term steers toward the desired path.
Notice the gain on~$\Delta_f$ is~$(k_f+1)$; the~$+1$ makes the pushing
angle~$\theta_p$ go beyond~$\Delta_f$ (with respect to~$\theta_d$), eventually
rotating the object back toward the desired pushing direction. Ultimately, the
controller converges to a configuration where the contact force points along
the desired path. The controller does not depend explicitly on any slider
parameters, and can thus be used to push a variety of unknown objects. Notably,
we do not require knowledge of the support friction, pressure distribution, or
contact friction, which are often uncertain and subject to change. Furthermore,
depending on the contact friction coefficient~$\mu_c$, the contact point may
slip or stick along the edge of the slider over the course of a successful
push.

In many cases we could just take~$\bm{v}_{\cmd}=\bm{v}_p$ and skip to
Sec.~\ref{sec:ik_ctrl}; however, there are a number of additions we can make to
our pushing controller to improve robustness and even handle collisions between
the slider and obstacles, which we discuss in the remainder of this section.

\subsection{Contact Recovery}

It is possible that following the pushing angle produced by~\eqref{eq:ctrl_law}
will cause the pusher to lose contact with the slider, especially with larger
gains~$k_f$ and~$k_c$. This can happen if the local curvature of the slider is
such that the angle~$\theta_p$ points away from the current contact edge.
Assuming quasistatic motion, we know that the slider does not move after
contact is broken. We will say that contact is lost if~$\|\bm{f}\|$ is less
than some threshold~$f_{\min}$. In the absence of a meaningful force
measurement with~$\|\bm{f}\|\geq f_{\min}$, a reasonable approach is to just
follow the desired path using the open-loop (with respect to the contact force)
control law
\begin{equation}\label{eq:ctrl_law_ol}
  \theta_o = \theta_d - k_c\Delta_c,
\end{equation}
which just steers the EE toward the desired path. Notice that the sign
of~$k_c\Delta_c$ is opposite to that in~\eqref{eq:ctrl_law}; this is because
here the pusher does not need to move \emph{away} from the path to steer the
slider back toward it.

Let's now combine~\eqref{eq:ctrl_law} and~\eqref{eq:ctrl_law_ol}.
Suppose that the pusher loses contact with the slider. Then our approach is to
rotate from the current pushing direction~$\theta_p$ toward the open-loop
angle~$\theta_o$ from~\eqref{eq:ctrl_law_ol}. When contact is made again, such
that~$\|\bm{f}\|\geq f_{\min}$, we switch back to~\eqref{eq:ctrl_law}. Thus
the combined EE velocity angle~$\theta_{\mathrm{ee}}$ is given by
\begin{equation}\label{eq:ctrl_law_with_convergence}
  \theta_{\mathrm{ee}} = \begin{cases}
    \theta_{\mathrm{ee}}^- + \gamma, & \text{if } \|\bm{f}\| < f_{\min} \\
    \theta_p, & \text{otherwise},
  \end{cases}
\end{equation}
where~$\theta_{\mathrm{ee}}^-$ is the value of~$\theta_{\mathrm{ee}}$ from the
previous control iteration and~$\gamma=\theta_o - \theta_{\mathrm{ee}}^-$ with
limit~$|\gamma|\leq\gamma_{\max}$. The corresponding EE velocity
is~$\bm{v}_{\mathrm{ee}}=v\,[\cos\theta_{\mathrm{ee}},\,\sin\theta_{\mathrm{ee}}]^T$.

We have also experimented with contact recovery mechanisms that attempts to
``circle back'' to the last contact point where~$\|\bm{f}\|\geq f_{\min}$. We
found~\eqref{eq:ctrl_law_with_convergence} to be somewhat more reliable in our
experiments, but a more sophisticated contact recovery mechanism is worth
investigating in future work.

\subsection{Obstacle Avoidance and Admittance Control}

\begin{figure}[t]
  \centering
    \includegraphics[width=0.8\columnwidth]{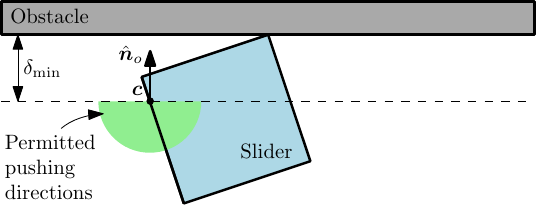}
    \caption{Basic obstacle avoidance of the pusher. When the contact
    point~$\bm{c}$ is within distance~$\delta_{\min}$ of an obstacle, the
    direction of motion is adjusted to not move any closer to it.}
  \label{fig:ee_obstacle}
\end{figure}

So far we have not said anything about obstacle avoidance. Consider the task of
pushing an object along a hallway. We will assume that the location of the
walls is known to the controller, so the pusher can avoid colliding with them.
To do this, if the EE is within some distance~$\delta_{\min}$ of an obstacle,
then we rotate~$\bm{v}_{\mathrm{ee}}$ by the smallest angle possible such that
it no longer points toward the obstacle (i.e., we
want~$\hat{\bm{n}}_o^T\bm{v}_{\mathrm{ee}}\leq0$, where~$\hat{\bm{n}}_o$ is the
unit vector pointing from the EE to the closest point on the obstacle; see
Fig.~\ref{fig:ee_obstacle}). However, since the geometry and pose of the slider
are not known, collisions between the slider and walls cannot be completely
avoided, especially if the hallway contains turns, so we need to handle these
collisions. It turns out that the pushing angle produced by~\eqref{eq:ctrl_law}
is still useful in many cases when the slider is in contact with obstacles.
However, we need to avoid producing excessively large forces by jamming the
slider against an obstacle, to prevent damage. We use an admittance controller
to adjust the velocity when~$\|\bm{f}\|$ is above a threshold~$f_{\max}$. In
particular, we compute a velocity offset
\begin{equation}\label{eq:admittance_ctrl}
  \bm{v}_a = \begin{cases}
    k_a(f_{\max}-\|\bm{f}\|)\skew5\hat{\bm{f}} & \text{if } \|\bm{f}\|>f_{\max}, \\
    \bm{0} & \text{otherwise,}
  \end{cases}
\end{equation}
where~$k_a>0$ is a tunable gain, and finally generate our commanded EE
velocity~$\bm{v}_{\mathrm{cmd}} = \bm{v}_{\mathrm{ee}} + \bm{v}_a$. The upshot
of this admittance control scheme is that the commanded
velocity~$\bm{v}_{\mathrm{\cmd}}$ is reduced in the direction of~$\bm{f}$
when~$\|\bm{f}\|$ is large, and can even move opposite to~$\bm{f}$. To avoid
excessive movement opposite~$\bm{f}$, we found it useful to clamp the magnitude
of~$\bm{v}_{\mathrm{cmd}}$ back to at most~$v$.

\subsection{Force Filtering}

The force measurements from our FT sensor are quite noisy, so we employ the
exponential smoothing filter
\begin{equation*}
  \bm{f}_{\mathrm{filt}} = \beta\bm{f}_{\mathrm{meas}} + (1-\beta)\bm{f}_{\mathrm{filt}}^-,
\end{equation*}
where~$\bm{f}_{\mathrm{filt}}$ is the filtered force, $\bm{f}_{\mathrm{meas}}$
is the raw measured force, and~$\beta=1-\exp(-\delta t/\tau)$ with the~$\delta
t$ the time between force measurements and~$\tau>0$ the tunable filter time
constant. We actually use this filtering approach in both simulation and
experiment; in simulation it helps to smooth out numerical noise in the
force values computed by the simulator. It should be assumed that all
references to~$\bm{f}$ elsewhere in the paper refer to the filtered value.

\section{Inverse Kinematics Controller}\label{sec:ik_ctrl}

The pushing controller described in the previous section produces a desired
velocity of the EE~$\bm{v}_{\mathrm{\cmd}}$ in task-space. We use an inverse kinematics (IK)
controller to realize the desired pushing velocity while avoiding collisions
between the robot body and known static obstacles. We use a planar
omnidirectional mobile robot with motion model~$\dot{\bm{q}} = \bm{u}$,
where~$\bm{q}=[x,y,\theta]^T$ is the robot's configuration, consisting of the
position~$(x,y)$ and the heading angle~$\theta$, and~$\bm{u}$ is the
corresponding joint velocity input. We use a quadratic program-based inverse
kinematics controller
\begin{equation}\label{eq:ik_ctrl}
  \begin{aligned}
    \bm{u} = \textstyle\argmin_{\bm{\xi}} &\quad (1/2)\|\bm{\xi}_d-\bm{\xi}\|^2 \\
    \text{subject to} &\quad \bm{J}_c(\bm{q})\bm{\xi} = \bm{v}_{\mathrm{cmd}} \\
                      &\quad {-}\bm{\xi}_{\max} \leq \bm{\xi} \leq \bm{\xi}_{\max},
  \end{aligned}
\end{equation}
where~$\bm{\xi}_d = [0, 0, k_{\omega}(\theta_d-\theta)]^T$ is designed to
minimize the linear velocity and the difference between the robot's
heading~$\theta$ and the path heading~$\theta_d$, with~$k_{\omega}>0$ a tunable
gain. The matrix~$\bm{J}_c(\bm{q})$ is the Jacobian of the contact point,
and~$\bm{\xi}_{\max}$ is the joint velocity limit. This IK controller allocates
the joint velocities such that the desired EE velocity is achieved exactly
while trading off between small linear velocities and rotating to match the
path's heading. In the presence of obstacles, we also add constraints to avoid
collisions with the robot base. We model the base as a circle with
center~$(x,y)$. If any part of this circle is within distance~$\delta_{\min}$
of an obstacle~$\mathcal{O}$, then we add the constraint~$[\hat{\bm{n}}_o^T,
0]\bm{\xi} \leq 0$ to~\eqref{eq:ik_ctrl}, where~$\hat{\bm{n}}_o$ is the unit
vector pointing from~$(x,y)$ to the closest point on~$\mathcal{O}$.

While here we have only considered a mobile robot with 3 degrees of freedom
(DOFs), which is sufficient to accomplish our pushing task, \eqref{eq:ik_ctrl}
has the structure of a standard IK controller and can be augmented with
additional DOFs, objectives, and constraints (see e.g.~\cite{heins2021mobile}),
as long as the required EE velocity for pushing is achieved.

\section{Simulation Experiments}

\begin{figure}[t]
  \centering
    \includegraphics[width=0.8\columnwidth]{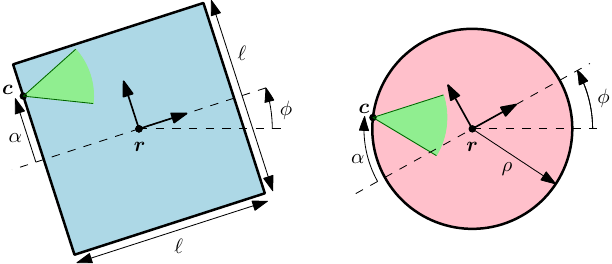}
    \caption{Examples of two sliders, located at
      position~$\bm{r}$ and orientation~$\phi$ in the global frame. The contact
      with the pusher is located at point~$\bm{c}$, which is distance~$\alpha$
      along the slider's edge from a reference point. The contact force must
      lie in the friction cone at the contact point (shown in green).}
  \label{fig:setup}
  \vspace{10pt}
\end{figure}

\begin{table}[t]
  \caption{Controller parameters for simulation and hardware experiments.}
  \centering
  \renewcommand{\arraystretch}{1.0}
  \begin{tabular}{c c c c}
    \toprule
    Parameter & Simulation & Hardware & Unit\\
    \midrule
    $v$ & 0.1 & 0.1 & \si{m/s} \\
    $k_f$ & 0.3 & 0.3 & -- \\
    $k_c$ & 0.1 & 0.5 & \si{rad/m} \\
    $k_a$ & 0.003 & 0.003 & \si{s/kg} \\
    $k_{\omega}$ & -- & 1 & $\si{1/s}$ \\
    $f_{\min}$ & 1 & 5 & \si{N} \\
    $f_{\max}$ & 50 & 50 & \si{N} \\
    $\gamma_{\max}$ & 0.1 & 0.1 & \si{rad} \\
    $\delta_{\min}$ & 0.1 & 0.1 & \si{m} \\
    $\tau$ & 0.05 & 0.05 & \si{s} \\
    $\bm{\xi}_{\max}$ & -- & $[0.5,0.5,0.25]^T$ & $[\si{m/s},\si{m/s},\si{rad/s}]^T$ \\
    \bottomrule
  \end{tabular}
  \label{tab:ctrl_parameters}
  \vspace{-5pt}
\end{table}

We first validate our controller in simulation with Box and Cylinder sliders
representing the planar sliders shown in Fig.~\ref{fig:setup}. We use the
PyBullet simulator\footnote{We applied a small patch to PyBullet to improve
sliding friction behaviour; see \texttt{\scriptsize
{https://github.com/bulletphysics/bullet3/pull/4539}}.}. The Box has $x$-$y$
side lengths~$\ell=\SI{1}{m}$ and height~\SI{12}{cm}; the Cylinder slider has
the same height and radius~$\rho=\SI{0.5}{m}$. Each has mass~$m=\SI{1}{kg}$
with CoM located at the centroid. The pusher is a sphere of radius~\SI{5}{cm}
and the height of the contact point is~\SI{6}{cm}. For each slider, we assess
the robustness of our controller by running simulations in different scenarios,
each of which has a different combination of lateral offset, contact offset,
slider orientation, contact friction, and slider inertia, as listed in
Table~\ref{tab:parameters}. We use the controller parameters listed in the
Simulation column of Table~\ref{tab:ctrl_parameters}. The simulation timestep
is~\SI{1}{ms} and the control timestep is~\SI{10}{ms} (i.e., the controller is
run once every 10 simulation steps). The friction coefficient between the
slider and floor and obstacles is set to~$\mu_o=0.25$. We also set the contact
stiffness and damping of the sliders to~$10^4$ and~$10^2$, respectively, for
stability during collisions between the slider and wall obstacles.\footnote{We
also performed simulations with~$\mu_o=0.5$ and with contact stiffness and
damping of~$10^5$ and~$10^3$, respectively, to ensure we could also handle
variation in these parameters. The results are similar to those shown here.}

\begin{table}[t]
  \caption{Initial states and parameters used for simulation. We refer to each combination
    of states and parameters as a scenario, for a total of~$3^5=243$ scenarios per
    slider. The values of the~$3\times3$ inertia matrix~$\bm{I}$ depend on the
    slider shape, with~$\skew4\hat{\bm{I}}$ computed assuming uniform density
    and~$\bm{I}_{\max}$ computed assuming all mass is concentrated in the outside
    wall of the Cylinder and in the vertices of the Box.}
  \centering
  \begin{tabular}{l c c c}
    \toprule
    Parameter & Symbol & Values & Unit \\
    \midrule
    Initial lateral offset & $\Delta_{c_0}$ & $-40$, $0$, $40$ & \si{cm} \\
    Initial contact offset & $\alpha_0$ & $-40$, $0$, $40$ & \si{cm} \\
    Initial orientation & $\phi_0$ & $-\pi/8$, $0$, $\pi/8$ & \si{rad} \\
    Contact friction & $\mu_c$ & $0$, $0.5$, $1.0$ & $-$ \\
    Slider inertia & $\bm{I}$ & $0.5\skew4\hat{\bm{I}}$, $\skew4\hat{\bm{I}}$, $\bm{I}_{\max}$ & $\si{kg{\cdot}m^2}$ \\
    \bottomrule
  \end{tabular}
  \label{tab:parameters}
\end{table}

\begin{figure}[t]
  \centering
    \includegraphics[width=\columnwidth]{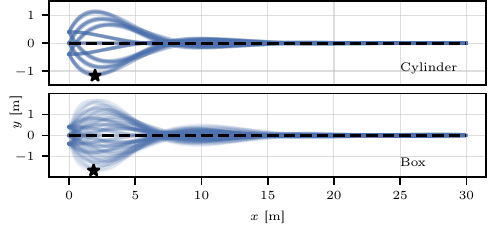}
    \caption{Simulated trajectories for all~$243$ scenarios
      given in Table~\ref{tab:parameters} for the Box and Cylinder sliders
      shown in Fig.~\ref{fig:setup}. Each trajectory has a
      duration of~\SI{5}{min}. All trajectories converge to the desired
      straight-line path using our control law. The point of maximum deviation
      from the path for any of the trajectories is marked with a star.}
  \label{fig:simulations_straight}
\end{figure}

\begin{figure}[t]
  \centering
    \includegraphics[width=\columnwidth]{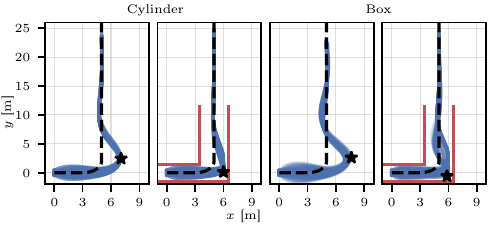}
    \caption{Simulated trajectories for the Box and Cylinder sliders
      along a curved path, with and without walls (in red) simulating a
      corridor. All trajectories again converge despite collisions with the
      wall. The point of maximum deviation
      from the path for any of the trajectories is marked with a star.}
  \label{fig:simulations_cor}
\end{figure}

The position trajectories for each of the~$3^5=243$ scenarios per slider are
shown in Fig.~\ref{fig:simulations_straight} with straight desired paths. Our
controller successfully steers both sliders to the desired path along the
positive $x$-axis for every scenario using the same controller parameters.
While~$k_c$ could be increased to reduce the deviation from the desired path,
we found that a larger~$k_c$ did not converge to a stable translational push
for all scenarios. The results of the same scenarios are shown in
Fig.~\ref{fig:simulations_cor} with a curved desired path, with and without
walls simulating a hallway corner. Without the walls, there is overshoot at the
turn before the slider ultimately returns to the desired path. With the walls,
the slider collides with the wall and the pushing velocity is adjusted by the
admittance controller~\eqref{eq:admittance_ctrl} before again eventually
returning to the desired path.

\begin{figure}[t]
  \centering
    \includegraphics[width=\columnwidth]{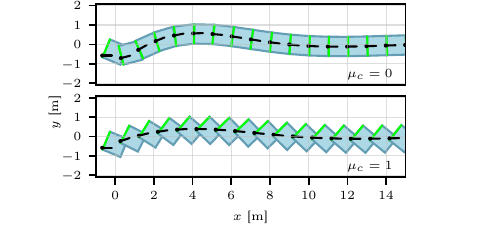}
    \caption{Samples of simulated trajectories of the Box slider with the
      straight desired path along the~$x$-axis. The slider has initial
      state~$(x_0,y_0,s_0,\phi_0)=(0,\SI{-40}{cm},\SI{-40}{cm},-\pi/8)$ and
      uniform density inertia. Results are shown for low and high contact
      friction. Pusher and pushing direction are shown in black, initial contact
      edge is highlighted in green. With~$\mu_c=0$, the contact point ultimately
      slides to the center of the contact edge; with~$\mu_c=1$, the contact point
      does not slide.}
  \label{fig:friction_example}
\end{figure}

\begin{figure}[t]
  \centering
  \includegraphics[width=0.42\columnwidth,valign=t,trim={0 0 0 0.08cm},clip]{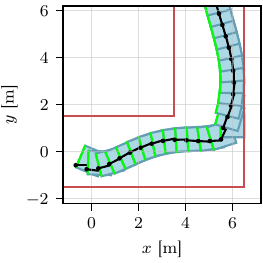}
    \hspace{0.25cm}
    \includegraphics[width=0.32\columnwidth,valign=t]{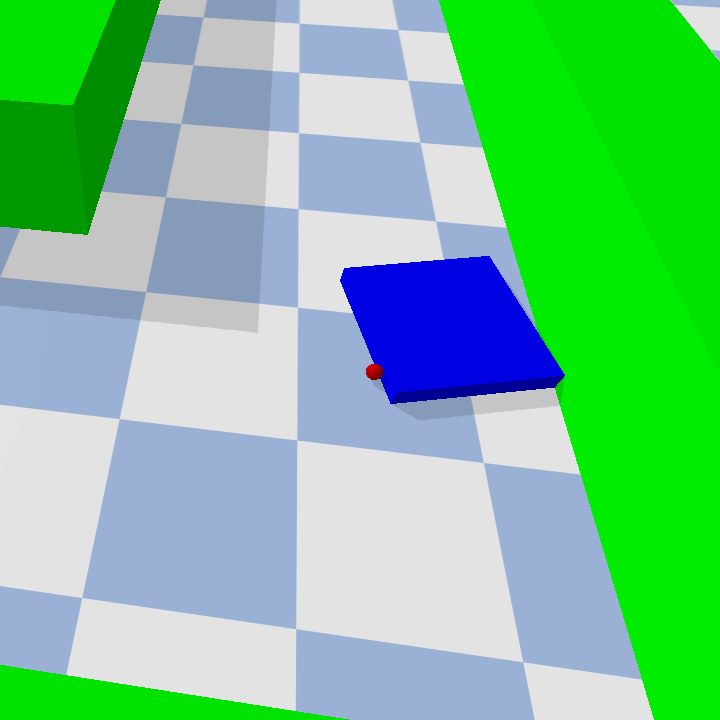}
    \caption{\emph{Left:} A sample simulated trajectory of the Box slider along the
      curved path with walls. The slider has initial
      state~$(x_0,y_0,s_0,\phi_0)=(0,\SI{-40}{cm},\SI{-40}{cm},-\pi/8)$, a
      uniform density inertia, and~$\mu_c=0$. Pusher and pushing
      direction are shown in black; the walls are red. The initial contact edge
      of the slider is highlighted in green. After contact with the
      wall, the pusher switches edges for the remainder of the trajectory.
      \emph{Right:} An image of the simulation, with red pusher,
      blue slider, and green walls. Readers are encouraged to watch the video
      to see this in more detail.}
  \label{fig:edge_switching_example}
\end{figure}

Individual sample trajectories are shown in Fig.~\ref{fig:friction_example} and
Fig.~\ref{fig:edge_switching_example}. In Fig.~\ref{fig:friction_example}, we
compare two trajectories along the straight-line path to demonstrate how the
behaviour of the system changes when sufficient friction to prevent slip at the
contact point is not available. The two scenarios are the same except that one
has no contact friction ($\mu_c=0$) and the other has high contact friction
($\mu_c=1$). With no contact friction, we see that the contact point quickly
slides to the middle of the contact edge. In contrast, with high friction, the
contact point does not slip and a stable translational push is achieved with a
large angle between the contact normal and pushing direction. In both cases,
the closed-loop system successfully converges to the desired path along
the~$x$-axis. In Fig.~\ref{fig:edge_switching_example}, we show a trajectory
along the curved path with walls. After colliding with the wall, the pusher
actually switches the contact edge\footnote{Technically it is a contact
\emph{face} since the sliders are three-dimensional objects, but we will say
edge given our planar context.} for the remainder of the trajectory. The
contact point slides along the original edge while attempting to turn the
slider, eventually briefly losing contact and circling back toward the path due
to~\eqref{eq:ctrl_law_with_convergence}, ultimately making contact again on a
different edge of the slider.

\section{Hardware Experiments}

\begin{figure}[t]
  \centering
  \includegraphics[width=0.48\columnwidth]{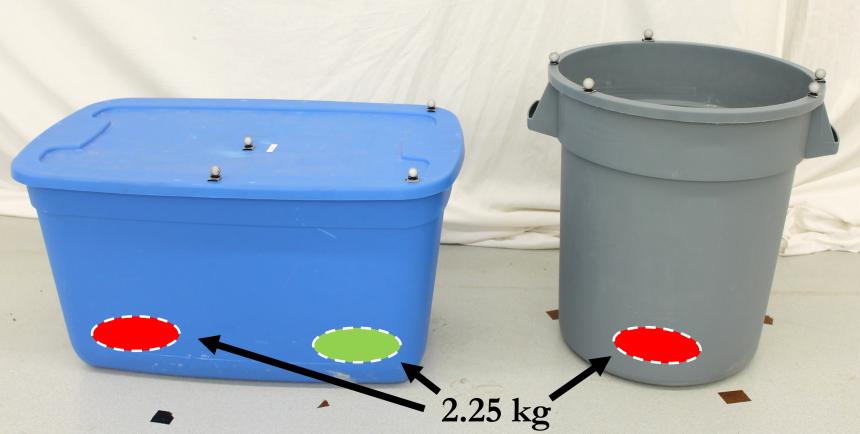}
  \includegraphics[width=0.48\columnwidth,trim={0 4cm 0 3cm},clip]{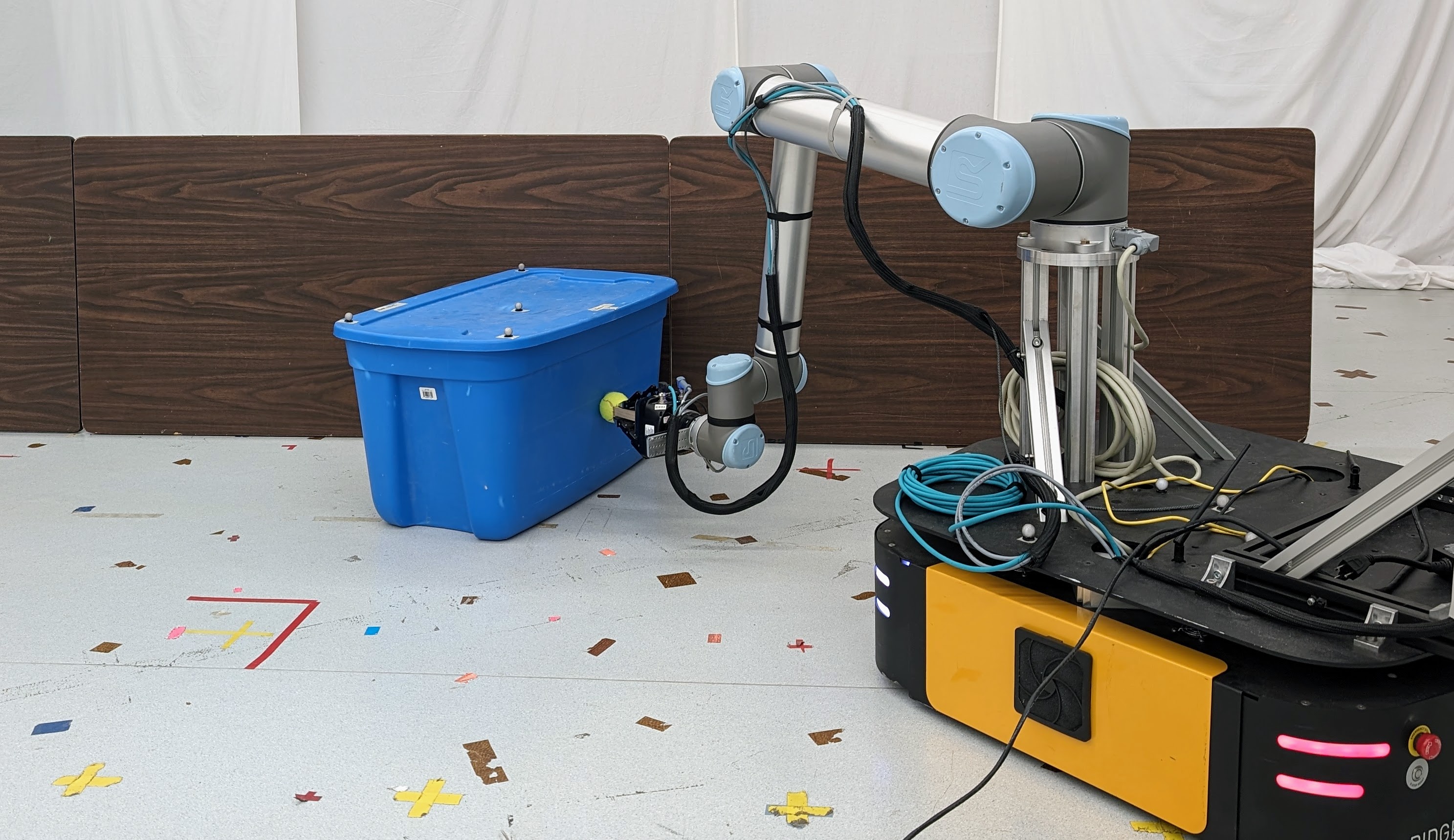}
  \caption{\emph{Left:} The ``Box'' and ``Barrel'' sliders used for real-world
      experiments. Each is empty except for~\SI{2.25}{kg} weights
      located approximately at the colored circles. The red weights are always
      present, but we add or remove the green weight to vary the mass and
      pressure distribution of the box. When both weights are present, we refer
      to the slider as ``Box2''; with a single weight it is called ``Box1''.
      The Barrel has radius~\SI{20.5}{cm}, height~\SI{56}{cm}, and total
      mass~\SI{4.5}{kg}. The Box has width~\SI{65}{cm}, depth~\SI{33}{cm}, and
      height~\SI{43}{cm}; the total mass of Box1 is~\SI{4.8}{kg} and of Box2
      is~\SI{7}{kg}. The friction coefficients with the ground were estimated
      to be approximately~$0.3$--$0.4$ on average for both objects.
      \emph{Right:} The robot pushing the Box along the curved trajectory,
      shortly after the slider first makes contact with the ``wall'' (we use
      overturned tables).}
  \label{fig:objects}
\end{figure}

We now demonstrate our controller in real-world experiments. The robot used for
pushing is a mobile manipulator consisting of a UR10 arm mounted on a Ridgeback
omnidirectional base (see Fig.~\ref{fig:eyecandy}). The arm's wrist is equipped
with a Robotiq FT 300 force-torque sensor. A tennis ball mounted at the EE is
used to contact the slider. Since we are only pushing in the
$x$-$y$ plane, we fix the joint angles of the arm and only control the base
using~\eqref{eq:ik_ctrl}---the arm is used only for the FT sensor. The
base is localized using a Vicon motion capture system that provides pose
measurements at~\SI{100}{Hz}, which is also used to record the trajectories of
the sliders (but the slider poses are \emph{not} provided to our controller).
The FT sensor provides force measurements at approximately~\SI{63}{Hz}, the
mobile base accepts commands at~\SI{25}{Hz}, and we run our control
loop\footnote{We could reduce the control frequency to match the slower command
  frequency of the robot, but our approach \emph{(a)} ensures the most up-to-date command
is sent to the robot, and \emph{(b)} demonstrates the efficiency of the controller.}
at~\SI{100}{Hz}. A video of the experiments can be found at \texttt{\small
\videourl}.

We test our controller's ability to push three sliders: Barrel, Box1, and Box2
(shown and described in~Fig.~\ref{fig:objects}). The height of the contact
point is constant and we assume it is low enough that the sliders do not tip
over. For each experiment, the slider starts slightly in front of the EE with
various lateral offsets; the robot moves forward until contact is made (i.e.,
$\|\bm{f}\|\geq f_{\min}$). For smoothness, the EE accelerates at a constant
rate over~\SI{1}{s} to reach the constant desired pushing speed~$v$. We use
ProxQP~\cite{bambade2022prox} to solve~\eqref{eq:ik_ctrl}. For obstacle
avoidance, we model the base as a circle of radius~\SI{55}{cm}. We use the
controller parameters in the Hardware column of
Table~\ref{tab:ctrl_parameters}, which are the same as in simulation except for
increased values of~$k_c$ and~$f_{\min}$. A lower value of~$k_c$ was required
in simulation so that the trajectory converged to the desired path for
\emph{all} combinations of parameters, but for these real-world experiments we
found that a higher~$k_c$ improves tracking performance. In general, the gains
can be tuned to give better tracking performance when the set of possible
slider parameters is smaller. We increased~$f_{\min}$ to reject noise in the
real-world FT sensor.

\begin{figure}[t]
  \centering
    \includegraphics[width=\columnwidth]{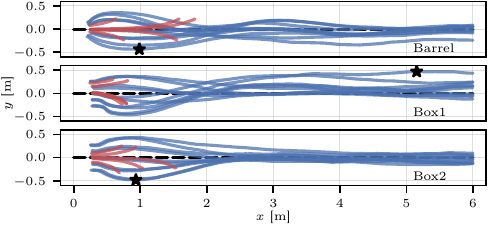}
    \caption{Position trajectories for real sliders pushed along a straight
      path starting from various lateral offsets. Ten trajectories using our
      pushing control law are shown for each slider (in blue), with the point
      of maximum deviation of any trajectory from the path marked by a star. We
      compare against an open-loop controller, which only tracks the path and
      does not use any slider feedback. Five open-loop trajectories are shown
      for each slider (in red). The open-loop trajectories end once contact
      between the EE and slider is lost. All open-loop trajectories fail within
      about~\SI{2}{m}, whereas our controller is able to push the objects
      across the full length of the room.}
  \label{fig:experiments_straight}
\end{figure}

The results for a straight-line desired path are shown in
Fig.~\ref{fig:experiments_straight}. Ten trajectories using our pushing control
law are shown for each slider. Here we compare against an open-loop controller,
which just follows the path using the open-loop angle~\eqref{eq:ctrl_law_ol}
and constant speed~$v$. Five open-loop trajectories are shown for each slider.
Open-loop pushing with single-point contact is not robust to changing friction,
misalignment with the slider's CoM, or other disturbances, and indeed we see
that the open-loop trajectories all fail within~\SI{2}{m} of the start of the
path, demonstrating the need for a closed-loop controller. In contrast, our
controller successfully pushes the sliders across the full~\SI{6}{m} length of
the room.

The trajectories do not converge perfectly to the desired path, at least not
within the available~\SI{6}{m} distance. This is expected in the real world, as
the slider is constantly perturbed by imperfections on the surface of the
ground as it slides, which must then be corrected by the controller. Indeed, as
can be seen in Fig.~\ref{fig:objects}, the floor of the room has various pieces
of tape and other markings which change the surface friction properties as the
object slides. For Box1, we actually expect the slider to end up slightly above
the path, since its CoM is offset from the measured position. Indeed, our
controller only ensures \emph{some} point on the slider (i.e., the contact
point) tracks the path, which depends on the slider's frictional and inertial
parameters. Regardless, in Fig.~\ref{fig:experiments_straight} we see that the
controller keeps the slider within approximately~\SI{0.5}{m} of the path at all
times, even with different pressure distributions and considerable initial
lateral offsets between pusher and slider, and converges to an even narrower
range.

\begin{figure}[t]
  \centering
    \includegraphics[width=2.75in]{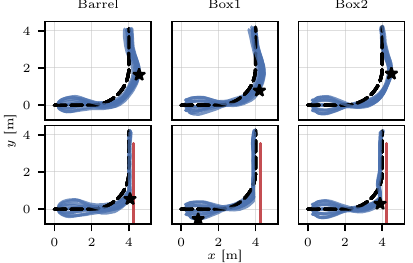}
    \caption{Position trajectories (in blue) for real sliders pushed along a
      curved path, using our pushing controller. In the top row, each slider is
      pushed through freespace; in the bottom row, an obstacle acting as a wall
      is introduced (in red), which blocks the motion of the sliders. Ten
      trajectories are shown for each scenario, with the point of maximum
      deviation from the path marked by a star. Despite hitting the
      wall, the controller adjusts the pushing velocity to continue
      pushing the sliders in the desired direction.}
  \label{fig:experiments_cor}
\end{figure}

The results for tracking a curved path are shown in
Fig.~\ref{fig:experiments_cor}. We show results for freespace as well as with
the addition of a wall, which blocks slider motion (see
Fig.~\ref{fig:objects}). The controller knows the location of the wall, so the
robot itself can avoid colliding with it. However, the slider does hit the
wall, after which the controller adjusts the pushing velocity to continue in
the desired direction. This setup represents a simplified version of navigating
a turn in a hallway, and demonstrates that we can, in principle, handle contact
with obstacles. In this work we assume the space is open enough that the robot
can always maneuver to obtain the desired EE velocity using~\eqref{eq:ik_ctrl};
future work will investigate narrower hallways and cluttered environments.

\begin{figure}[t]
  \centering
    \includegraphics[width=\columnwidth]{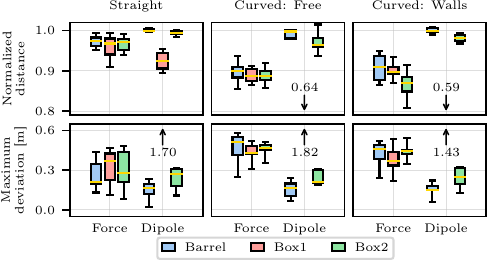}
    \caption{Boxplots of metrics for our proposed approach (``Force''; 10
      trajectories per boxplot) and the vision-based baseline
      from~\cite{igarashi2010a} (``Dipole''; 5 trajectories per boxplot). The
      middle yellow line is the median, the box represents the first and third
      quartiles, and the whiskers represent the minimum and maximum values. The
      numbered arrows indicate the medians of values outside the axis limits.
      The normalized distance is the distance actually travelled along the path
      divided by the distance that would have been travelled if the path were
      perfectly tracked; the maximum deviation is the farthest point of the
      slider from the desired path. The normalized distance can exceed 1 if
      velocity tracking is imperfect or if some of the path is skipped, but the
      latter necessarily results in some path deviation. Notice that while the
      dipole approach is effective when the slider's position is well-aligned
      with its CoM, it deviates substantially if not (e.g., with Box1).}
  \label{fig:experiment_stats}
\end{figure}

Finally, we compare our force-based controller to a vision-based controller
(i.e., one that uses measurements of the slider's position, which we obtain
using motion capture). We use the ``dipole'' approach
from~\cite{igarashi2010a}, which generates pushing directions based on the
measured angle between the desired goal position and the EE position with
respect to the slider. The goal position is the point~\SI{1}{m} ahead of the
current closest point on the desired path. Fig.~\ref{fig:experiment_stats}
presents metrics comparing the controllers. The maximum deviation metric is
simply the farthest distance between the slider and the desired path, which
gives a measure of worst-case path-tracking error. The normalized distance
metric is calculated as follows. Let~$t_0$ be the time of first contact,
let~$t_f$ be the final time, and let~$\bm{p}_{d_0}$ and~$\bm{p}_{d_f}$ be the
closest points on the path to the slider's position at~$t_0$ and~$t_f$,
respectively. Then we define the ideal distance travelled
as~$\bar{d}=v(t_f-t_0)$ and the actual distance travelled~$d$ as the distance
along the path between~$\bm{p}_{d_0}$ and~$\bm{p}_{d_f}$. The normalized
distance~$d/\bar{d}$ is a measure of how well the task was completed relative
to an ``ideal'' controller that tracked the path perfectly. For simplicity, we
neglect the short acceleration phase at the start of each trajectory.

\begin{figure}[t]
  \centering
    \includegraphics[width=\columnwidth]{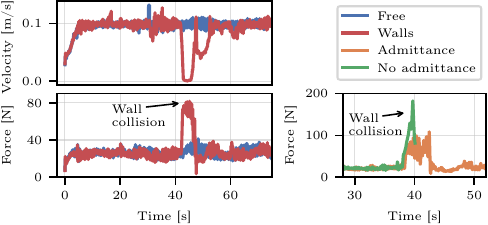}
    \caption{\emph{Left:} Slider velocity and contact force magnitudes from two
      runs of the Box2 slider along the curved path, with and without the wall
      obstacle. The force increases and the velocity decreases upon collision
      with the wall, until the controller adjusts the pushing velocity to
      continue along the path. \emph{Right:} Contact force magnitudes from two
      different runs of the Box2 slider along the curved path with walls, with
      and without the admittance controller. Without the admittance controller,
      the maximum force is well in excess of~\SI{150}{N}, and the experiment
      was stopped to avoid damage. The other run has the highest maximum force of
      any runs that used the admittance controller, which is much lower than~\SI{150}{N}.}
  \label{fig:vel_force_example}
\end{figure}

Looking at Fig.~\ref{fig:experiment_stats}, let us first examine our proposed
force-based approach. We see that the normalized distance is higher and the
maximum deviation is lower for the straight path compared to the curved
one---the curved path is in some sense more difficult. The metrics between each
of the sliders are fairly similar for a given desired path. The addition of the
wall also does not substantially alter the metrics for the curved path, though
the normalized distance for Box2 is slightly lowered. The wall briefly slows
down the slider after collision, but this removes the overshoot from the path
that occurs when the wall is not present (see Fig.~\ref{fig:vel_force_example}
for an example of the change in contact force and slider velocity that occurs
when colliding with the wall). Second, consider the dipole approach. We expect
a vision-based approach to generally outperform one using only the contact
force, since the measured force is noisy and only provides local information at
the contact point. Indeed, the dipole approach is effective when the measured
position is closely aligned with the slider's CoM, but results in large errors
when it is not (e.g., with Box1). This would require extra online adaptation to
resolve, something which our force-based controller provides automatically.
Ultimately, one must decide which approach (or a combination) makes sense given
the available sensing infrastructure---but now force-based pushing is a
possible option. Overall, our force-based controller is able to efficiently
navigate the path while keeping the deviation reasonable, and we encourage
readers to gain more insight into the controller's behaviour by watching the
supplemental video.

\section{Conclusion}

We presented a control law for quasistatic robotic planar pushing with
single-point contact using force feedback, which does not require known slider
parameters or slider pose feedback. We demonstrated its robustness in simulated
and real-world experiments, including collisions with a static wall obstacle,
which show that our controller reliably converges to the desired path with
reasonable deviation errors. Future work includes a formal proof of stability,
more sophisticated force-based controllers that can improve performance over
time through interaction with the slider, and investigation of hybrid
approaches that combine force and vision.

\bibliographystyle{IEEEtran}
\bibliography{bibliography}

\end{document}